\documentclass[conference, letterpaper]{ieeeconf}
\IEEEoverridecommandlockouts
\usepackage{cite}
\usepackage{amsmath,amssymb,amsfonts}
\usepackage{graphicx}
\usepackage{textcomp}
\usepackage{xcolor}
\def\BibTeX{{\rm B\kern-.05em{\sc i\kern-.025em b}\kern-.08em
    T\kern-.1667em\lower.7ex\hbox{E}\kern-.125emX}}

\usepackage{algorithm}
\usepackage{algorithmicx}  
\usepackage[noend]{algpseudocode}
\usepackage{amsfonts}
\usepackage{amsmath}
\usepackage{amsthm}
\usepackage{bm}
\usepackage{booktabs}
\usepackage{diagbox}
\usepackage{dsfont} 
\let\labelindent\relax  
\usepackage{enumitem}
\usepackage{etoolbox}
\usepackage{graphicx}  
\usepackage{hyperref}  
\usepackage{lipsum}  
\usepackage{mathtools}
\usepackage{multirow}
\usepackage{nicefrac}
\usepackage[final]{pdfpages}
\usepackage{siunitx}
\usepackage[caption=false]{subfig}
\usepackage{tabu}
\usepackage{threeparttable}  
\usepackage{tikz}
\usepackage{todonotes}
\usepackage{xcolor}
\usepackage{xfrac}

\usepackage{pgfplots}   
\pgfplotsset{compat=newest}
\usepgfplotslibrary{groupplots}
\usepgfplotslibrary{fillbetween}
\usetikzlibrary{pgfplots.statistics} 
\usetikzlibrary{arrows.meta}

\usepackage{mathabx}   

\usepackage{marvosym}  
\graphicspath{{img/}}
\hypersetup{colorlinks=false}

\definecolor{githubColor}{HTML}{2EA44F}
\newcommand{\gitref}[2]{\href{#1}{\color{githubColor}{#2}}}%

\definecolor{newGray}{HTML}{808080}

\definecolor{backGray}{rgb}{0.3, 0.3, 0.3}%
\definecolor{matlabYellow}{rgb}{0.9290, 0.6940, 0.1250}%
\definecolor{matlabPurple}{rgb}{0.4940, 0.1840, 0.5560}%
\definecolor{matlabLBlue}{rgb}{0.3010, 0.7450, 0.9330}%
\definecolor{matlabGreen}{rgb}{0.4660, 0.6740, 0.1880}%
\definecolor{matlabRed}{rgb}{0.8500, 0.3250, 0.0980}%
\definecolor{matlabBlue}{rgb}{0, 0.4470, 0.7410}%
\definecolor{matlabDarkRed}{rgb}{0.6350 0.0780 0.1840}%

\definecolor{colorCircle}{HTML}{0072BD}

\definecolor{colorRect}{HTML}{D95319}

\newcolumntype{O}[1]{S[detect-weight, mode=text, table-format=#1]}

\renewcommand{\bfseries}{\fontseries{b}\selectfont} 
\robustify\bfseries             
\newrobustcmd{\B}{\bfseries}

\newcommand\copyrighttext{\footnotesize \textcopyright~2024 IEEE. Personal use of this material is permitted. Permission from IEEE must be obtained for all other uses, in any current or future media, including reprinting/republishing this material for advertising or promotional purposes, creating new collective works, for resale or redistribution to servers or lists, or reuse of any copyrighted component of this work in other works.
}

\newcommand\copyrightnotice{%
    \begin{tikzpicture}[remember picture,overlay]%
 	\node[anchor=south, xshift=-0pt, yshift=20pt] at (current page.south)%
 	{\fbox{\parbox{\dimexpr\textwidth-\fboxsep-\fboxrule\relax}{\copyrighttext}}};%
 	\end{tikzpicture}%
}

\newtheoremstyle{tstyle}
  {}
  {}
  {\itshape}
  {}
  {\bfseries}
  {.}
  { }
  {\thmname{#1}\thmnumber{ #2}\thmnote{ (#3)}}%
\theoremstyle{tstyle}

\newcommand*\diff{\mathop{}\!\mathrm{d}}

\newcommand{\inv}{^\text{\rmfamily \textup{-1}}}

\newcommand{\argmin}{\operatorname{arg\,min}}

\newcommand{\pspace}{\,}  

\colorlet{sensorblue}{matlabBlue!70!white}
\colorlet{sensorbluew}{matlabBlue!40!white}
\colorlet{sensorred}{matlabRed!70!white}
\colorlet{sensorredw}{matlabRed!40!white}
\colorlet{sensorgreen}{matlabGreen!70!white}
\colorlet{sensorgreenw}{matlabGreen!40!white}
\colorlet{sensorgray}{newGray!60!white}
\colorlet{sensorgrayw}{newGray!30!white}

\tikzset{bluefade/.style={top color=sensorblue, bottom color=sensorbluew}}
\tikzset{redfade/.style={top color=sensorred, bottom color=sensorredw}}
\tikzset{greenfade/.style={top color=sensorgreen, bottom color=sensorgreenw}}
\tikzset{grayfade/.style={top color=sensorgray, bottom color=sensorgrayw}}

\title{%
Self-Assessment and Correction of Sensor Synchronization%
}

\author{Thomas Wodtko, Alexander Scheible  and Michael Buchholz%
\thanks{Parts of this work were supported by the State Ministry of Economic Affairs, Labour and Tourism Baden-Württemberg (project U-Shift\,II, AZ\,3-433.62-DLR/60).%
Parts of this research have been conducted as part of the PoDIUM project, which is funded by the European Union under grant agreement No. 101069547. Views and opinions expressed are, however, those of the authors only and do not necessarily reflect those of the European Union or European Commission. Neither the European Union nor the granting authority can be held responsible for them.}%
\thanks{All authors are with the Institute of Measurement, Control, and Microtechnology, Ulm University, Albert-Einstein-Allee 41, 89081 Ulm, Germany {\tt\footnotesize \{firstname\}.\{lastname\}@uni-ulm.de}}%
}

\begin{document}

\maketitle

\begin{abstract}
We propose an approach to assess the synchronization of rigidly mounted sensors based on their rotational motion. 
Using function similarity measures combined with a sliding window approach, our approach is capable of estimating time-varying time offsets.
Further, the estimated offset allows the correction of erroneously assigned time stamps on measurements.
This mitigates the effect of synchronization issues on subsequent modules in autonomous software stacks, such as tracking systems that heavily rely on accurate measurement time stamps.
Additionally, a self-assessment based on an uncertainty measure is derived, and correction strategies are described.
Our approach is evaluated with Monte Carlo experiments containing different error patterns.
The results show that our approach accurately estimates time offsets and, thus, is able to detect and assess synchronization issues.
To further embrace the importance of our approach for autonomous systems, we investigate the effect of synchronization inconsistencies in tracking systems in more detail and demonstrate the beneficial effect of our proposed offset correction.
\end{abstract}


\section{Introduction}
\copyrightnotice%
\label{sec:intro}
Synchronization is vital for automated systems.
Sensor data acquisition and processing, as well as communications, must be in sync to guarantee proper operation.
For example, time stamp deviations of acquired sensor data deteriorate the performance of the consecutive fusion step~\cite{jellum2024sync}.
Different methods are available to synchronize sensors and compute units inside autonomous systems~\cite{jellum2024sync}, such as, e.g., IEEE1588 precision time protocol (PTP)~\cite{ieee1588}.
However, due to the lack of verification approaches, successful synchronization is often only assumed during operation.

Recent regulatory advances for the operation of autonomous vehicles proposed functional safety standards such as the ISO 21448 safety of the intended functionality (SOTIF)~\cite{iso21448}.
A key aspect of this standard's requirements is the development of self-assessment modules, such as for the tracking of objects.
While approaches are available for some of these modules~\cite{griebel2023online}, they only assess specific functionalities.
Thus, any synchronization issues would only indirectly be recognized with a delay.
Accordingly, assessing the synchronization of all automated system parts is desirable.
To the best of our knowledge, no approach is currently available for this.

In this work, assuming rigidly connected sensors, we will leverage properties already known in the field of motion-based extrinsic calibration of sensors~\cite{daniilidis1999handeye,horn2021online}.
There, sensor ego-motion estimates are used to yield the transformation between two sensor frames, which is called extrinsic calibration.
As described later, it can be shown that the rotational motion of rigidly connected sensors must be the same at all times.
This property was already used by the authors of~\cite{taylor2016motion}~\&~\cite{furrer2017Evaluation} to synchronize a constant time offset during calibration.
Extending its applicability, we will use the same property and expand the estimation of a single offset during calibration to the estimation of a time-dependent and non-constant offset during operation.

In the following, we describe related methods and required fundamentals in detail in Section~\ref{sec:relWork}~\&~\ref{sec:foundations}, respectively.
Then, our proposed approach is derived in Section~\ref{sec:method} and evaluated based on experiments in Section~\ref{sec:experiments}.

\begin{figure}[t]
    \centering
    \resizebox{0.95\linewidth}{!}{%
\begin{tikzpicture}[baseline=(current bounding box.center), double distance=2pt]

\def\recw{2}
\def\rech{1}
\def\recoff{0.75}

\def\ws{0}
\def\wocor{8}
\def\wpro{11}

\def\hstruc{4}
\def\struch{3}
\def\strucmargin{0.5}

\def\estw{3}

\def\avMar{0.3}
\def\ahMar{0.75}
\def\alMar{0.125}

\shadedraw [bluefade] (\ws,0) rectangle ++(\recw,-\rech) node [midway] {Sensor1};
\shadedraw [redfade] (\ws,-\rech-\recoff) rectangle ++(\recw,-\rech) node [midway] {Sensor2};

\shadedraw [bluefade]  (\wocor,0) rectangle ++(\recw,-\rech) node [midway, align=center] {Timestamp\\Correction};
\shadedraw [redfade]  (\wocor,-\rech-\recoff) rectangle ++(\recw,-\rech) node [midway, align=center] {Timestamp\\Correction};

\shadedraw [grayfade] (\wpro,0) rectangle ++(\recw,-2*\rech-\recoff) node [midway] {Processing};

\shadedraw [greenfade] (\ws+\strucmargin,-\hstruc) rectangle ++(\wocor-\ws+\recw-2*\strucmargin,-\struch);

\draw [rounded corners=0.5cm] (\ws+2*\strucmargin ,-\hstruc-\strucmargin) rectangle ++(\estw,-\struch+2*\strucmargin) node [midway, align=center] {Offset\\Estimation};
\draw [rounded corners=0.5cm] (\wocor+\recw-2*\strucmargin-\estw ,-\hstruc-\strucmargin) rectangle ++(\estw,-\struch+2*\strucmargin) node [midway, align=center] {Self-Assessment\\\& Correction};

\draw [-{Latex[round, scale=1.5]}] (\ws+\recw+\alMar, 0-\avMar) -- (\wocor-\alMar, 0-\avMar) node [midway,yshift=0.25cm] {Sensor Data};
\draw [-{Latex[round, scale=1.5]}] (\ws+\recw+\alMar, -\rech-\recoff-\avMar) -- (\wocor-\alMar, -\rech-\recoff-\avMar) node [midway,yshift=0.25cm] {Sensor Data};

\draw [-{Latex[round, scale=1.5]}] (\wocor+\recw+\alMar, 0-0.5*\rech) -- (\wpro-\alMar, 0-0.5*\rech);
\draw [-{Latex[round, scale=1.5]}] (\wocor+\recw+\alMar, -\rech-\recoff-0.5*\rech) -- (\wpro-\alMar, -\rech-\recoff-0.5*\rech);
   
\draw [draw=sensorblue, -{Latex[round, scale=1.5]}] 
    (\ws+\recw+\alMar, -\rech+\avMar) -- 
    ++(2*\strucmargin+\estw-\ahMar-\recw-\alMar,0) -- 
    (\ws+2*\strucmargin+\estw-\ahMar, -\hstruc-\strucmargin+\alMar);   
\node[color=sensorblue, rotate=-90] (V1) at (\ws+2*\strucmargin+\estw-\ahMar+0.25, -\hstruc-\strucmargin+\alMar + 1.1) {Motion};

\draw [draw=sensorred, -{Latex[round, scale=1.5]}] 
    (\ws+\recw+\alMar, -2*\rech-\recoff+\avMar) -- 
    ++(2*\strucmargin+\estw-2*\ahMar-\recw-\alMar,0) -- 
    (\ws+2*\strucmargin+\estw-2*\ahMar, -\hstruc-\strucmargin+\alMar) ;   
\node[color=sensorred, rotate=-90] (V2) at (\ws+2*\strucmargin+\estw-2*\ahMar+0.25, -\hstruc-\strucmargin+\alMar + 1.1) {Motion};

\draw [draw=sensorblue, {Latex[round, scale=1.5]}-] 
    (\wocor-\alMar, -\rech+\avMar) -- 
    ++(\recw-2*\strucmargin-\estw+\alMar+\ahMar,0) -- 
    (\wocor+\recw-2*\strucmargin-\estw+\ahMar, -\hstruc-\strucmargin+\alMar);   
\node[color=sensorblue, rotate=90] (C1) at (\wocor+\recw-2*\strucmargin-\estw+\ahMar-0.25, -\hstruc-\strucmargin+\alMar + 1.3) {Correction}; 

\draw [draw=sensorred, {Latex[round, scale=1.5]}-] 
    (\wocor-\alMar, -2*\rech-\recoff+\avMar) -- 
    ++(\recw-2*\strucmargin-\estw+\alMar+2*\ahMar,0) -- 
    (\wocor+\recw-2*\strucmargin-\estw+2*\ahMar, -\hstruc-\strucmargin+\alMar);   
\node[color=sensorred, rotate=90] (C1) at (\wocor+\recw-2*\strucmargin-\estw+2*\ahMar-0.25, -\hstruc-\strucmargin+\alMar + 1.3) {Correction};

\draw [-{Latex[round, scale=1.5]}] 
    (\ws+2*\strucmargin+\estw+\alMar, -\hstruc-\strucmargin-2*\avMar) -- 
        (\wocor+\recw-2*\strucmargin-\estw-\alMar, -\hstruc-\strucmargin-2*\avMar) 
        node [midway,yshift=0.25cm] {Offset};
\draw [-{Latex[round, scale=1.5]}] 
    (\ws+2*\strucmargin+\estw+\alMar, -\hstruc-\struch+\strucmargin+2*\avMar) -- 
    (\wocor+\recw-2*\strucmargin-\estw-\alMar, -\hstruc-\struch+\strucmargin+2*\avMar) 
    node [midway,yshift=0.25cm] {Uncertainty};

\draw [-{Latex[round, scale=1.5]}] 
    (\wocor+\recw-2*\strucmargin+\alMar, -\hstruc-0.5*\struch) --
    (\wpro+\recw-4*\alMar, -\hstruc-0.5*\struch) 
    node [midway,yshift=0.25cm] {Self-Assessment};

\end{tikzpicture}
}
    \caption{%
        The overall structure of our proposed self-assessment and correction approach is illustrated.
        It can mainly be divided into two parts, the offset estimation and the self-assessment and correction part.
        First, the offset estimation uses motion estimates of two sensors.
        A possible timestamp offset and a respective estimation uncertainty are estimated based on the rotational movements.
        Second, the self-assessment investigates the estimated offset over time.
        When deviations are detected, the self-assessment additionally decides, based on the uncertainty, whether timestamps can be corrected or data needs to be discarded.
        Respectively, sensor data is updated or, if necessary, discarded in the timestamp correction block before other modules further process it.
    }
    \label{fig:SNRain}
    \vspace{-0.5cm}
\end{figure}
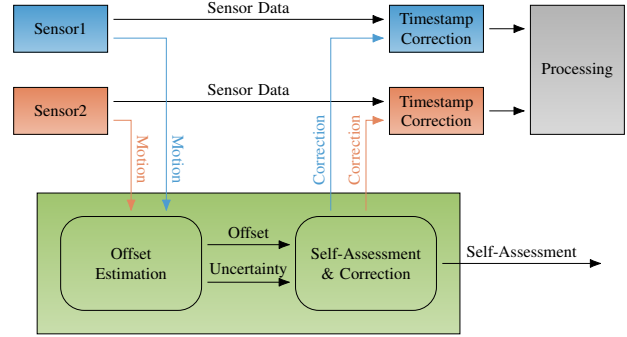

Summarizing our work in this paper, we propose
\begin{itemize}[topsep=0pt]
    \item an self-assessment and correction approach for the synchronization of sensors,
    \item an impact analysis of synchronization offsets in tracking systems, and
    \item an open-source extension for the open-source library Excalibur~\cite{excalibur}\footnote{\gitref{https://github.com/uulm-mrm/excalibur}{https://github.com/uulm-mrm/excalibur}} (upon publication).
\end{itemize}

\section{Related Work}
\label{sec:relWork}
First, this section describes the importance of sensor synchronization and self-assessment in detail.
Then, available synchronization approaches in the field of extrinsic sensor calibration are presented and related to our proposed approach.

\subsection{Synchronization in autonomous systems}
A variety of methods is available to synchronize sensors and processing units of autonomous systems for both hardware and software~\cite{jellum2024sync}.
Thereby, it can mostly be distinguished between two strategies.
First, a trigger signal can be used to initiate the sensor data acquisition.
Second, each component of the autonomous system has an independent clock, and all clocks are synchronized.
For the latter, the data acquisition is triggered by each sensor separately based on the internal clock.
In either case, time stamps are assigned to the acquired data, which are of importance in later steps~\cite{jellum2024sync}.
At this point, in some specific settings for some approaches, the actual synchronization of components may be verified, e.g. when all components are directly connected, and IEEE~1588~PTP~\cite{ieee1588} is used.
However, this is not possible for most approaches, and the correct assignment of timestamps to sensor data is assumed for later processing.
To the best of our knowledge, there is no approach available that allows online verification at any point after a time stamp was assigned.
Thus, any synchronization issue entails misbehavior in later modules, e.g. in sensor fusion~\cite{jellum2024sync}, where the time of acquisition is used to align measurements.
While self-assessment approaches for some processing modules exist~\cite{griebel2023online}, only general misbehavior can be detected in this case, and specific conclusions about, e.g., the synchronization of data are not possible.
Here, our work fills the gap in self-assessment modules for sensor synchronization.

\subsection{Time Offset Estimation in Motion-based Calibration}
In motion-based extrinsic calibration, motion estimates of sensors are used to obtain the extrinsic calibration of sensors.
The authors of~\cite{daniilidis1999handeye} showed that the rotation of rigidly connected sensors is the same at any time and excluded it from the calibration process.
However, when sensors are synchronized, we demonstrated in previous work~\cite{horn2021online} that considering the rotation difference can enhance the calibration result due to noise.
This shows that noise will lead to discrepancies even for synchronized sensor data.
In contrast, the authors of~\cite{taylor2016motion, furrer2017Evaluation} use the rotational motion to synchronize motion estimated prior to calibration.
In~\cite{taylor2016motion}, a likelihood function considering uncertainties is created specifically for the time offset and is included in the optimization process.
Here, the estimation works well for initially small time offsets and large data sets.
Smaller data sets lead to higher errors.
In ~\cite{furrer2017Evaluation}, the cross-correlation function is used to obtain the time offset between two sensors; no investigation of different noise levels or data set sizes is available.
Both approaches yield a constant time offset, meaning it is assumed that both sensors acquired data synchronously and only the recorded timestamp is incorrect.
In this work, however, we assume that two sensors lose their synchronization, which, in general, can lead to drifting or stepping offsets for which the two mentioned approaches are not suitable.
In order to allow changes in the offset, a sliding window will be used.
Here, the window size relates to the data set length in~\cite{taylor2016motion}.

To the best of our knowledge, no approach is available to assess the synchronization of sensors and detect changing time offsets online.

\section{Foundations}
\label{sec:foundations}

Before our method is derived, some fundamentals are described in this section.

\subsection{Motion in 3D Space}
\begin{figure}[t]
    \centering
    \newcommand{\drawSensors}[3]{%
\begin{scope}[shift={#1}, rotate=#2]

    \draw[draw=sensorgray, fill=sensorgrayw, rounded corners=0.5cm] (-0.5,1.5) rectangle (0.5, -1.5);
    \node[circle, bluefade,transform shape] (A#3) at (0,1) {$A_{#3}$};
    \node[circle, redfade,transform shape] (B#3) at (0,-1) {$B_{#3}$};    

    \draw[-{Latex[round, scale=1.5]}, transform shape] (B#3) -- (A#3) node [midway, xshift=0.25cm] {$T$};
\end{scope}
}

\resizebox{0.95\linewidth}{!}{%
\begin{tikzpicture}[baseline=(current bounding box.center), double distance=2pt, shorten >= 2pt, shorten <= 2pt]

\drawSensors{(0,0)}{0}{1};
\drawSensors{(2.5,1)}{10}{2};
\drawSensors{(5,0.5)}{-10}{3};
\drawSensors{(7.5,0)}{10}{4};

\draw[draw=sensorblue,-{Latex[round, scale=1.5]}] (A1) -- (A2) node [sensorblue, midway, sloped, yshift=0.3cm] {$V_a^1$};
\draw[draw=sensorblue,-{Latex[round, scale=1.5]}] (A2) -- (A3) node [sensorblue, midway, sloped, yshift=0.3cm] {$V_a^2$};
\draw[draw=sensorblue,-{Latex[round, scale=1.5]}] (A3) -- (A4) node [sensorblue, midway, sloped, yshift=0.3cm] {$V_a^3$};

\draw[draw=sensorred,-{Latex[round, scale=1.5]}] (B1) -- (B2) node [sensorred, midway, sloped, yshift=0.3cm] {$V_b^1$};
\draw[draw=sensorred,-{Latex[round, scale=1.5]}] (B2) -- (B3) node [sensorred, midway, sloped, yshift=0.3cm] {$V_b^2$};
\draw[draw=sensorred,-{Latex[round, scale=1.5]}] (B3) -- (B4) node [sensorred, midway, sloped, yshift=0.3cm] {$V_b^3$};

\node[scale=2, newGray] at (1.25,0.5) {$\circlearrowleft$};
\node[scale=2, newGray] at (3.75,0.75) {$\circlearrowleft$};
\node[scale=2, newGray] at (6.25,0.25) {$\circlearrowleft$};

\end{tikzpicture}%
}
    \caption{%
        The general motion of two sensors over time is shown. 
        Sensor 1 (blue) and sensor 2 (red) are rigidly connected with the transformation $T$.
        $A_i$ and $B_i$ represent different positions of sensor 1 and sensor 2 at time step $i$, respectively;
        and $V^A_i$ and $V^B_i$ represent the motion of a sensor between two consecutive time steps.
        For each motion, the transformation cycle is marked with a gray circle.
    }
    \label{fig:motionGraph}
    \vspace{-0.5cm}
\end{figure}
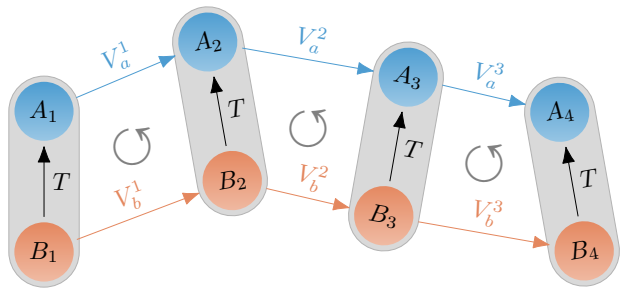

In the following, transformations possess six degrees of freedom (DoF) and are generally referred to as a function; thus, no specific implementation is implicitly specified.
A motion graph can describe the motion of objects in 3D space.
An example of two moving and rigidly connected sensors is illustrated in Fig.~\ref{fig:motionGraph}.
Due to physical reasons, each transformation cycle in Fig.~\ref{fig:motionGraph} must be closed, i.e. the concatenation of all transformations of the cycle must result in an identity transformation.
For a single time step, this is defined by
\begin{align}
\label{eq:transcycle}
    T \circ V_a \circ T\inv \circ V_b\inv = I \pspace ,
\end{align}
where $I$ denotes the identity transformation.
Using, e.g., dual quaternions, it can easily be shown that the rotation magnitude of $V_a$ and $V_b$ must equal to hold  Eq.~\eqref{eq:transcycle}~\cite{daniilidis1999handeye}.
This means that when two rigidly connected sensors are moving, and their motion is synchronously measured, the measured rotation magnitude is the same.
As a consequence, assuming non-periodic and non-constant rotation, the rotation magnitude of sensor motion can be considered as an indicator for synchronous motion measurements.
A time shift of rotation magnitudes would indicate a shift in the measurement acquisition time.
This property is used in~\cite{furrer2017Evaluation} to estimate a constant time offset between sensor measurements.
In Section~\ref{sec:method}, we will also leverage this property to derive a method for changing time offset.

\subsection{Similarity of Functions}
Later in this work, two functions need to be compared; for this, a similarity measure is required.
Generally, the cross-correlation of two functions~\cite{bracewell1996fourier} is a similarity measure based on convolution.
Given the two functions $f: \mathbb{R} \mapsto \mathbb{R}$ and $g: \mathbb{R} \mapsto \mathbb{R}$, the cross-correlations function $\phi: \mathbb{R} \mapsto \mathbb{R}$ is defined by
\begin{align}
    \phi(\tau) = \int_{-\infty}^{\infty} f(t) \cdot g(t+\tau) \pspace \diff t \pspace .
\end{align}
For two functions $f, g\colon I\mapsto \mathbb{R}$ defined on the finite interval $I = [0 \, \dots \, N-1]\subset \mathbb{N}_0$ with $N\in\mathbb{N}$ elements, the finite time-discrete cross correlation function $\phi: I \mapsto \mathbb{R}$ is defined by
\begin{align}
    \phi[\tau] = \sum_{m = 0}^{N - 1} f[m] \cdot g[(m+\tau)_{\textrm{mod} \, N} ] \pspace .
\end{align}
Here, using the modulo operator, the periodicity of functions is assumed.
Otherwise, the value of the cross-correlation function would automatically decrease with an increasing $\tau$ since fewer and fewer discrete steps overlap.
The cross-correlation function yields a similarity score for the two functions given a shift of $\tau$.
Meaning that all possible shifts between the two functions are evaluated.
The absolute score value depends on the scale of the two functions.

While the periodicity assumption is often valid in the field of signal processing, it is not suitable for the later task of this work.
Respectively, we derive a different similarity measure with a similar structure later in Section~\ref{sec:method}.

\subsection{Kalman Filtering}
\begin{figure}
    \centering
    \includegraphics[width=\linewidth,trim={0.5cm 26.5cm 12.5cm 1cm},clip]{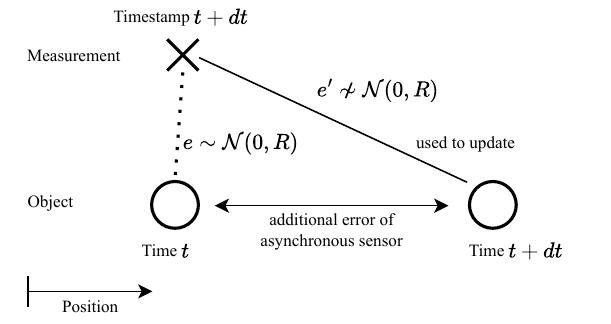}
    \caption{
        The effect of a measurement time offset on the tracking system is illustrated.
        A moving object (circle) is shown at different times.
        At time $t$, the object gets measured by a sensor (cross), but due to synchronization issues, the sensor incorrectly assigns the time step $t+dt$ to the measurement.
        Now, the tracking system would use this measurement to update the estimation of time $t+dt$. 
        This introduces additional errors and violates the assumptions of the Kalman filter.
    }
    \label{fig:effect}
\end{figure}

Later, we demonstrate the effect of synchronous sensors on autonomous vehicles using the tracking module as an example.
Here, the Kalman filter~\cite{kalman1960new} is the most basic filter that tracks a single object.
The filter consists of two steps: the prediction and the update step. 
During the prediction, the estimation $X_t$ of time $t$ gets predicted to the next time \hbox{$t+\Delta t$} using the process model.
The Kalman filter assumes a Gaussian distributed state and a linear process model with Gaussian noise.
Let
\begin{align}
    X_{t, t}\sim \mathcal{N}(\mu_{t, t}, \Sigma_{t, t})
\end{align}
be the estimation at time point $t$. 
Then, the predicted state at time $t'=t+\Delta t$ is given by 
\begin{align}
    X_{t', t} \sim \mathcal{N}(F_{\Delta t}\mu_{t, t}, F_{\Delta t} \Sigma_{t, t} F_{\Delta t}^T + Q_{\Delta t}),
\end{align}
where $F_{\Delta t}$ denotes the process matrix and $Q_{\Delta t}$ the covariance of the noise.
The specific form of $F$ and $Q$ depends on the process model.
In this work, we use a two-dimensional nearly constant velocity model with 
\begin{subequations}
    \begin{align}
    F_{\Delta t} &= \begin{pmatrix}
        1 & \Delta t \\
        0 & 1        \\
        1 & \Delta t \\
        0 & 1 \\
    \end{pmatrix}, \\[1ex]
    Q_{\Delta t} &= \Gamma \sigma \Gamma^T \qquad \text{with} \\[1ex]
    \Gamma &= \begin{pmatrix}
        \Delta t^2 / 2 & 0 \\
        \Delta t & 0       \\
        0 & \Delta t^2 / 2 \\
        0 & \Delta t       \\
    \end{pmatrix} \\[1ex]
    \sigma &= \text{diag}(\sigma^2_{a_x},~\sigma^2_{a_y}).
\end{align}%
\end{subequations}
In the update step, the measurement $z_{t'}$ is used to correct the predicted estimation.
The Kalman filter assumes a linear measurement model given by the measurement matrix $H$ and Gaussian noise with zero-mean, so the measurement follows 
\begin{subequations}
\begin{align}
  z_{t'} &= Hx_{t'} + e \\
  e&\sim\mathcal{N}(0, R), \label{eq:meas}
\end{align}    
\end{subequations}
where $x_{t'}$ is the true object state at time $t'$ and $R$ is the covariance of the measurement.
Then, the updated state \hbox{$X_{t',t'}\sim\mathcal{N}(\mu_{t', t'}, \Sigma_{t', t'})$} is also Gaussian distributed with 
\begin{subequations}
    \begin{align}
    K &= \Sigma_{t', t}H^T(H \Sigma_{t', t} H^T + R) \\
    \mu_{t', t'} &= \mu_{t', t} + K (z_{t'} - H \mu_{t', t}) \\
    \Sigma_{t', t'} &= (I - KH) \Sigma_{t', t}
\end{align}
\end{subequations}
An asynchronous sensor with a time offset $dt$ affects the filter by introducing a bias in the measurements because it updates the state with the wrong measurement, as illustrated in Fig.~\ref{fig:effect}.
This means the measurement error $e$ no longer fulfills (\ref{eq:meas}), and the filter performance decreases.

\section{Problem Formulation}
In this work, we assume that sensor data is acquired together with a timestamp.
Further, motion transformation estimates are available for each data point.
Given a set of \hbox{$n\in\mathbb{N}$} sensors $S = \{s_1, \dots, s_n\}$, motion estimates $V_k^i\in SE(3)$ at each timestamp $k\in\mathbb{N}$ and sensor $i\in\mathbb{N}$, the goal is to estimate a time offset $t_k\in\mathbb{R}$ for each timestamp.
Generally, the time offset may change over time.
Additionally, an uncertainty measure should be estimated based on the idea that the time offset error should be small if the uncertainty is small.
It shall be noted, that high uncertainty measures need not indicate large time offset estimation errors.

\section{Method}
\label{sec:method}

In this section, we derive our approach in two steps.
First, the time offset between two sensors is derived, and then the self-assessment and correction are described.
Given the motion estimates $V_k^i$, only the rotational part is considered in the following.
For this, the rotation magnitude of $V_k^i$ is denoted by $r_k^i$.

\subsection{Time Offset Estimation}
Wlog starting at time step $k_0=0$, the rotation magnitude of a sensor $i$ can be described by the time-discrete function $r_i: \mathbb{N}_0 \mapsto \mathbb{R}_+$.
Assuming that the time offset does not change over time, the offset between two sensors can be estimated using the cross-correlation function, as shown in ~\cite{furrer2017Evaluation}, and the argument of its maximum indicates the time offset.
Extending this to changing time offsets, we first consider a window of the last $w$ rotations estimates at any time.
Let $W = [0 \, \dots \, w-1]$ be the discrete window interval and the finite, time-discrete function $r_{i,k}: W \mapsto \mathbb{R}; \, r_{i,k}[l] := r_i[k-w+l+1]$ represent the window of rotation magnitudes of sensor $i$ at time step $k$.
Next, the similarity of this window for two different sensors must be considered.
First, the window is linearly interpolated with a factor of $b$ to increase precision.
Now, with $\widecheck{w} = w \cdot b$ and the interpolated window interval $\widecheck{W} = [0 \, \dots \, \widecheck{w} - 1]$, the interpolated rotation magnitude $\widecheck{r}_{i,k}: \widecheck{W} \mapsto \mathbb{R}$ contains $b$ sample points between each time step of the original function.

Since we cannot assume periodicity but the scale of values must be equal, we consider the sum of the absolute differences $\theta: \widecheck{W}_s \mapsto \mathbb{R}_+$ with the shifted window interval $\widecheck{W}_s = [-\frac{\widecheck{w}}{2} \, \dots \, \frac{\widecheck{w}}{2}-1]$ instead of the cross-correlation function.
For two rotation magnitude windows $\widecheck{r}_{1,k}$ and $\widecheck{r}_{2,k}$ it is defined by
\begin{align}
\label{eq:origthet}
    \theta[s] = \begin{cases}
        \sum\limits_{m = 0}^{\widecheck{w}-1-s} \Bigl\lvert \widecheck{r}^s_{\text{diff}}[m] \Bigr\rvert \quad & \text{if } s \geq 0 \\
        \sum\limits_{m = -s}^{\widecheck{w}-1} \Bigl\lvert \widecheck{r}^s_{\text{diff}}[m] \Bigr\rvert & \text{else}
    \end{cases} \pspace ,
\end{align}
with $\widecheck{r}^s_{\text{diff}}: [0 \dots \widecheck{w}-1-s] \mapsto \mathbb{R} ; \, \widecheck{r}_{\text{diff}} := \widecheck{r}_{1,k}[m] - \widecheck{r}_{2,k}[m + s]$.
While cross correlation function maps the indices to ensure a constant function overlap, the amount of overlapping samples is reduced in Eq.~\eqref{eq:origthet} with an increasing shift $s$.
Further, all samples inside the window are equally weighted, but increasing the influence of more recent samples seems desirable.
To overcome these issues, we introduce correction terms to the similarity measure.
First, the sum is normalized by the number of overlapping samples using a normalization function $\eta: \widecheck{W}_s \mapsto \mathbb{R}$ defined by
\begin{align}
    \eta[s] &= \frac{1}{w - |s|}
\end{align}
Second, a temporal correction function $\tau: \widecheck{W} \mapsto \mathbb{R}$ increases weights of more recent samples.
It is defined as
\begin{align}
    \tau[m] &= \overline{\tau}^{\,m / \widecheck{w}} \pspace ,
\end{align}
where $\overline{\tau}\in(0, 1)$ is a temporal factor that can be used to adjust the progression of weight decay over time.
The similarity measure $\theta$ from Eq.~\eqref{eq:origthet} is extended with the additional terms leading to the extended similarity measure $\widecheck{\theta}: \widecheck{W}_s \mapsto \mathbb{R}$ defined by
\begin{align}
\label{eq:extthet}
    \widecheck{\theta}[s] = \begin{cases}
        \eta[s]\sum\limits_{m = 0}^{\widecheck{w}-1-s} \tau[m] \, \Bigl\lvert \widecheck{r}^s_{\text{diff}}[m] \Bigr\rvert \quad & \text{if } s \geq 0 \\
        \eta[s]\sum\limits_{m = -s}^{\widecheck{w}-1} \tau[m+s] \, \Bigl\lvert \widecheck{r}^s_{\text{diff}}[m] \Bigr\rvert & \text{else}
    \end{cases} \pspace .
\end{align}
Now, the argument of the minimum of the extended similarity measure indicates the estimated time offset $s^*$. 
Formally, the time offset estimation is defined as
\begin{align}
    s^* = \frac{1}{b} \, \argmin_{s \in \widecheck{W}} \, \widecheck{\theta}[s] \pspace .
\end{align}

Next, given the estimation, an uncertainty measure is desirable that assesses the quality of the estimated time offset.
More specifically, whenever the uncertainty is low, the respective time offset estimation error must be small; however, when the uncertainty is high, only the error assessment is not possible, meaning that the error need not be large.
The estimation above requires certain characteristics in the rotational motion in order to successfully obtain the time offset, i.e., without a change of the rotation magnitude, no offset can be estimated.
The more rotational changes are within the considered window, the better suitable the similarity measure used by the offset estimation.
Respectively, the sum of absolute rotational changes is considered a certainty measure.
Formulary, the uncertainty measure $u: \mathbb{N}_0 \mapsto \mathbb{R}_+$ is defined by
\begin{align}
    u[k] = 1 / \left( \sum\limits_{i = 1} ^ 2 \sum\limits_{l=0}^{w-2} \Bigl\lvert r_{i,k}[l+1] - r_{i,k}[l]  \Bigr\rvert \right) \pspace .
\end{align}

\subsection{Self-Assessment \& Correction}
Based on the time offset estimation and its uncertainty measure, the self-assessment aims at assessing the overall performance.
Further, if applicable, the sensor data time stamps should be corrected by the estimated offset.
Generally, the self-assessment is interface-specific, meaning that depending on the expected output format, the available information can be transformed.
In this work, the self-assessment shall evaluate the performance of the time offset estimation only.
Thus, the uncertainty measure directly reflects the self-assessment.
However, depending on the system at hand, additional aspects might be of interest and should be evaluated, e.g. the state of synchronicity between sensors.  
When integrating our approach into a self-assessment framework considering a broader range of modules, it seems desirable to generate subjective logic (SL) opinions to represent the self-assessment states, similar to~\cite{griebel2023online}.
Here, multiple aspects, e.g. the state of synchronicity and the estimation performance, can potentially be considered differently.
However, it would go beyond the scope of this work and is open to future work.

Correcting time offsets is vital to maintain system integrity in the event of synchronization issues.
Different correction strategies are described next.
Focusing on the time offset estimation, a trivial correction strategy is to apply every estimated offset to its respective sensor measurement. 
Here, only the current step has to be considered; however, in case of uncertain measurements, potential errors in the offset estimation might decrease the system performance.
Instead, the uncertainty measure can be considered.
Corrections can be applied only when the estimation performance is assumed appropriate and the uncertainty is below a certain threshold.
Other measurements can simply be discarded.
However, as later shown in Section~\ref{sec:experiments}, depending on the vehicle motion, many measurements would get discarded this way, which in turn would decrease the system performance as well.
Assuming that any correction is beneficial in cases of high offsets, estimated offsets can only be applied when exceeding a certain offset threshold or when the uncertainty is below a threshold, but no measurement is discarded.
Here, it is assumed that any correction is beneficial in the case of high offsets, but small offsets might have less impact on the system performance than incorrect estimations.
All described strategies have both advantages and disadvantages; choosing a suitable correction strategy is task-specific.
For example, when multiple sensors are available, discarding a measurement seems more suitable than considering incorrect offset estimations.
Further, with more information about the system at hand, other correction strategies are possible.

\section{Experiments}
\label{sec:experiments}
In this section, we present the results of our evaluation based on simulated experiments.
If not differently stated, all results are averaged over $1\,000$ simulation runs.
All evaluations were run on a computer containing an ADM\,Ryzen\textsuperscript{TM}\,7\,3700X CPU, and 64GB of DDR4 RAM.

For two sensors mounted on an ego vehicle and for a target vehicle, we simulated data, as described in Section~\ref{sec:dataSim}.
Further, we simulated each sensor's measurement of the target vehicle at each step.
In Section~\ref{sec:offsetEst}, we present the results of the offset estimation in two different scenarios.
Then, in Section~\ref{sec:trackeval}, we analyze the impact of time offsets on information fusion in tracking systems and present the improvements using our proposed method.

\subsection{Data Simulation}
\label{sec:dataSim}
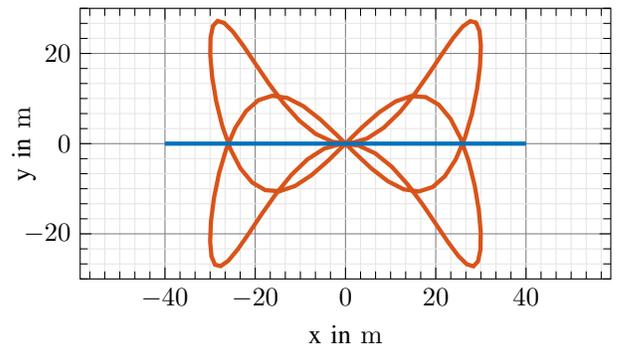
\begin{figure}
    \centering
    \vspace{0.2cm}
    \begin{tikzpicture}
\begin{axis}[
    axis equal,
    width = 1.0 \linewidth,
    height = 0.6 \linewidth,
    tick style={color=black},
    ymin=-30,
    ymax=30,
    xminorticks=true,
    yminorticks=true,
    minor x tick num=5,
    minor y tick num=5,
    xmajorgrids,
    xminorgrids,
    ymajorgrids,
    yminorgrids,
    major grid style={black!50},
    minor grid style={black!10},
    xlabel = x in \si{\metre},
    ylabel = y in \si{\metre},
    ylabel style={yshift=-4mm, xshift=-0mm},
]

    \newcommand\radius{30.0}
    \newcommand\sinFac{2.0}
    \newcommand\cosFac{3.0}

    \addplot [matlabRed, samples=100, domain=0:2*pi, ultra thick] 
    (%
        {\radius * sin(\sinFac * deg(x))}, %
        {\radius * sin(\sinFac * deg(x)) * cos(\cosFac * deg(x)}%
    );
    \addplot [matlabBlue, samples=200, domain=-40:40, ultra thick] 
    (%
        {x}, %
        {0}%
    );
\end{axis}
\end{tikzpicture} 
    \caption{
        The simulated path of the ego vehicle (red) and the target vehicle (blue).
        Both move on a plane, resulting in 2D motion.
    }
    \label{fig:simulationPath}
\end{figure}
Data is simulated similarly to our previous work~\cite{horn2021online}.
Thereby, a 2D path is projected onto a 3D surface, illustrated in Fig.~\ref{fig:simulationPath}.
We selected a plane surface for simplicity and illustration reasons, meaning that all simulated motion prior to noise is 2D only.
Further, unbiased Gaussian noise is added to each motion.
Its standard derivation, called noise level, is set with relative values.
For example, if the average motion along the path rotates $\SI{0.5}{\radian}$, a noise level of $\SI{10}{\percent}$ leads to Gaussian noise with $\sigma_{\text{rot}} = \SI{0.05}{\radian}$.
To improve simulation, motion is simulated on a grid $100$ times finer than it is used later on, i.e. when $200$ time steps are required, $20\,000$ steps are simulated instead, noise is added, and shifts are realized before the resulting motion is batched to yield the required $200$ motion simulations.
This allows us to realize time offsets with a higher resolution than the sensor motion is sampled.
For example, given the above values, time offsets of $0.01$ or $0.15$ are possible where otherwise the offset would have to be an integer value. 
Given the position of both the ego and target vehicle, sensor measurements are simulated for each sensor at every time step.

\subsection{Offset Estimation}
\label{sec:offsetEst}
\begin{figure*}
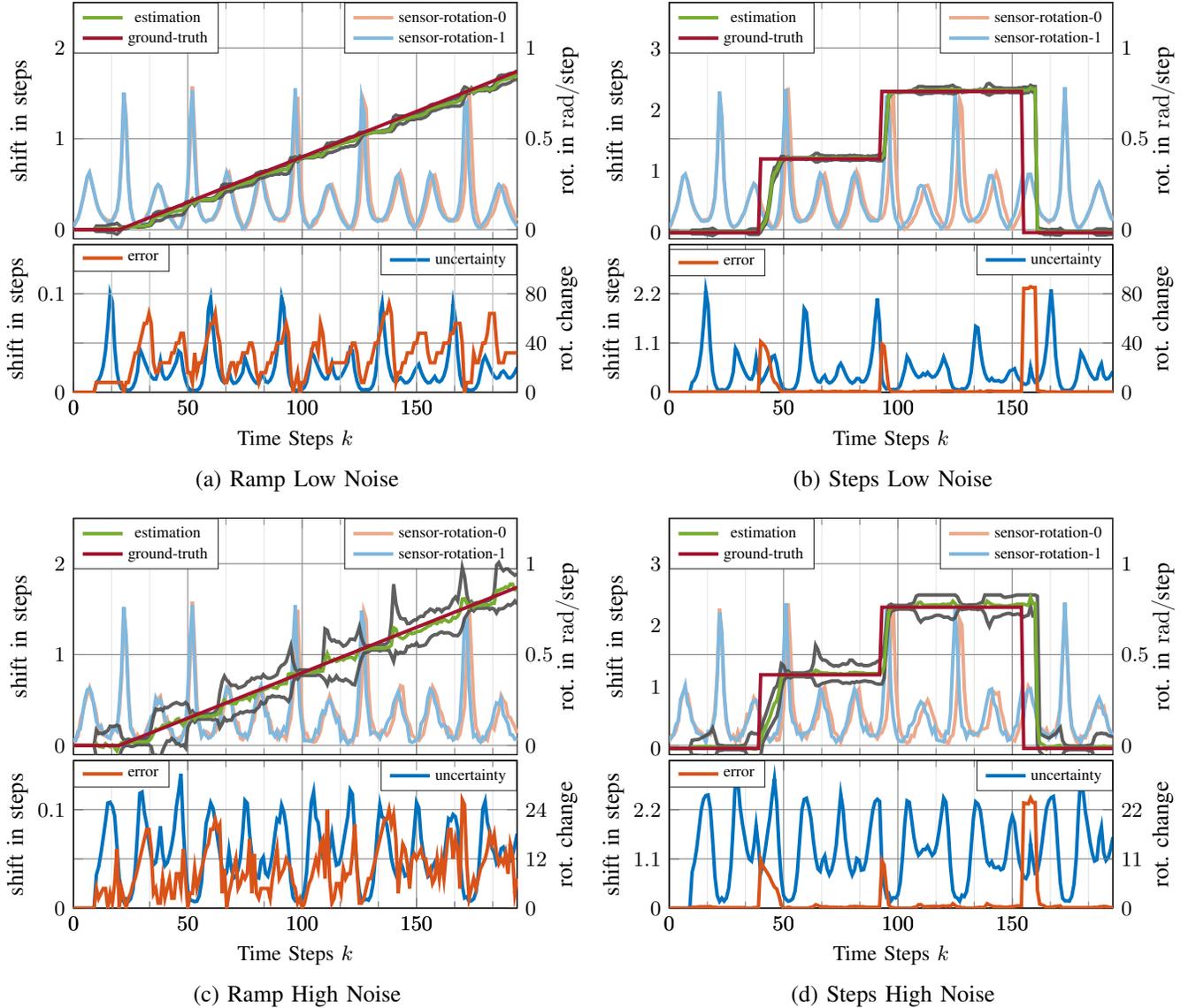

    \centering
    \captionsetup[subfloat]{labelfont=normalsize,textfont=normalsize}
    \subfloat[Ramp Low Noise]{%
       \begin{tikzpicture}[baseline=(current bounding box.center)]
    
    \colorlet{sensor-rotation-0Color}{matlabRed!50!white}
    \colorlet{ground-truthColor}{matlabDarkRed}
    \colorlet{sensor-rotation-1Color}{matlabBlue!50!white}
    \colorlet{estimationColor}{matlabGreen}
    
    \colorlet{errorColor}{matlabRed}
    \colorlet{uncertaintyColor}{matlabBlue}

    \begin{groupplot}[
        group style={
            group size=1 by 2,
            group name=ramp,
            x descriptions at=edge bottom,
            y descriptions at=edge right,
            vertical sep=2.5pt,
            xlabels at=edge bottom
        },
        legend cell align={left},
        xlabel style={font=\color{white!15!black}, font=\small},
        xlabel={Time Steps $k$},
        minor x tick num=2,
        width=0.5 * 0.75\linewidth,
        scale only axis,
        xmin=0,
        xmax=194,
        xminorticks=true,
        axis background/.style={fill=white},
        xmajorgrids,
        xminorgrids,
        ymajorgrids,
        yminorgrids,
        major grid style={black!50},
        minor grid style={black!10},
        ticklabel style = {font=\small},
    ]
    
    \nextgroupplot[
        height = 0.5 * 0.4\linewidth,
        ymin=-0.05,
        ymax=1.25,
        yminorticks=true,
        ylabel style={yshift=2mm, xshift=-0mm},
        ylabel={rot. in $\si{\radian} / \text{step}$},
        legend columns = 1,
        legend style={at={(axis cs: 194,1.25)},anchor=north east, font = \small, legend cell align=center, align=center, draw=white!15!black,nodes={scale=0.75, transform shape}}
    ]

    \input{img/raw/exp/ramp_low_rotation}

    \nextgroupplot[
        height = 0.5 * 0.25\linewidth,
        ymin=-0.0,
        ymax=120,
        yminorticks=true,
        ytick={0, 40, 80},
        ylabel style={yshift=1mm, xshift=-0mm},
        ylabel={rot. change},
        legend columns = 1,
        legend style={at={(axis cs: 194.,120.)},anchor=north east, font = \small, legend cell align=center, align=center, draw=white!15!black,nodes={scale=0.75, transform shape}}
    ]

    \input{img/raw/exp/ramp_low_uncertainty}
    
    \end{groupplot}

    \begin{groupplot}[
        group style={
            group size=1 by 2,
            group name=ramp,
            x descriptions at=edge bottom,
            y descriptions at=edge left,
            vertical sep=2.5pt,
            xlabels at=edge bottom
        },
        legend cell align={left},
        minor x tick num=2,
        width=0.5 * 0.75\linewidth,
        scale only axis,
        xmin=0,
        xmax=194,
        ticklabel style = {font=\small},
        xmajorgrids,
        xminorgrids,
        ymajorgrids,
        yminorgrids,
        xminorticks=true,
    ]
    
    \nextgroupplot[
        height = 0.5 * 0.4\linewidth,
        ymin=-0.1,
        ymax=2.5,
        yminorticks=true,
        ylabel style={yshift=1mm, xshift=-0mm},
        ylabel={shift in steps},
        legend columns = 1,
        legend style={at={(axis cs: 0.,2.5)},anchor=north west, font = \small, legend cell align=center, align=center, draw=white!15!black,nodes={scale=0.75, transform shape}}
    ]

    \input{img/raw/exp/ramp_low_estimation}

    \nextgroupplot[
        height = 0.5 * 0.25\linewidth,
        ymin=-0.0,
        ymax=0.15,
        yminorticks=true,
        ytick={0, 0.1},
        ylabel style={yshift=-1mm, xshift=-0mm},
        ylabel={shift in steps},
        legend columns = 1,
        legend style={at={(axis cs: 0.,0.15)},anchor=north west, font = \small, legend cell align=center, align=center, draw=white!15!black,nodes={scale=0.75, transform shape}}
    ]

    \input{img/raw/exp/ramp_low_error}
    
    \end{groupplot}
    
\end{tikzpicture}%
    }
    \subfloat[Steps Low Noise]{%
       \begin{tikzpicture}[baseline=(current bounding box.center)]
    
    \colorlet{sensor-rotation-0Color}{matlabRed!50!white}
    \colorlet{ground-truthColor}{matlabDarkRed}
    \colorlet{sensor-rotation-1Color}{matlabBlue!50!white}
    \colorlet{estimationColor}{matlabGreen}
    
    \colorlet{errorColor}{matlabRed}
    \colorlet{uncertaintyColor}{matlabBlue}

    \begin{groupplot}[
        group style={
            group size=1 by 2,
            group name=steps,
            x descriptions at=edge bottom,
            y descriptions at=edge right,
            vertical sep=2.5pt,
            xlabels at=edge bottom
        },
        legend cell align={left},
        xlabel style={font=\color{white!15!black}, font=\small},
        xlabel={Time Steps $k$},
        minor x tick num=2,
        width=0.5 * 0.75\linewidth,
        scale only axis,
        xmin=0,
        xmax=194,
        xminorticks=true,
        axis background/.style={fill=white},
        xmajorgrids,
        xminorgrids,
        ymajorgrids,
        yminorgrids,
        major grid style={black!50},
        minor grid style={black!10},
        ticklabel style = {font=\small},
    ]
    
    \nextgroupplot[
        height = 0.5 * 0.4\linewidth,
        ymin=-0.05,
        ymax=1.25,
        yminorticks=true,
        ylabel style={yshift=2mm, xshift=-0mm},
        ylabel={rot. in $\si{\radian} / \text{step}$},
        legend columns = 1,
        legend style={at={(axis cs: 194,1.25)},anchor=north east, font = \small, legend cell align=center, align=center, draw=white!15!black,nodes={scale=0.75, transform shape}}
    ]

    \input{img/raw/exp/steps_low_rotation}

    \nextgroupplot[
        height = 0.5 * 0.25\linewidth,
        ymin=-0.0,
        ymax=120,
        yminorticks=true,
        ytick={0, 40, 80},
        ylabel style={yshift=1mm, xshift=-0mm},
        ylabel={rot. change},
        legend columns = 1,
        legend style={at={(axis cs: 194.,120.)},anchor=north east, font = \small, legend cell align=center, align=center, draw=white!15!black,nodes={scale=0.75, transform shape}}
    ]

    \input{img/raw/exp/steps_low_uncertainty}
    
    \end{groupplot}

    \begin{groupplot}[
        group style={
            group size=1 by 2,
            group name=steps,
            x descriptions at=edge bottom,
            y descriptions at=edge left,
            vertical sep=2.5pt,
            xlabels at=edge bottom
        },
        legend cell align={left},
        minor x tick num=2,
        width=0.5 * 0.75\linewidth,
        scale only axis,
        xmin=0,
        xmax=194,
        ticklabel style = {font=\small},
    ]
    
    \nextgroupplot[
        height = 0.5 * 0.4\linewidth,
        ymin=-0.1,
        ymax=3.75,
        yminorticks=true,
        ylabel style={yshift=1mm, xshift=-0mm},
        ylabel={shift in steps},
        legend columns = 1,
        legend style={at={(axis cs: 0.,3.75)},anchor=north west, font = \small, legend cell align=center, align=center, draw=white!15!black,nodes={scale=0.75, transform shape}}
    ]

    \input{img/raw/exp/steps_low_estimation}

    \nextgroupplot[
        height = 0.5 * 0.25\linewidth,
        ymin=-0.0,
        ymax=3.3,
        ytick={0, 1.1, 2.2},
        ylabel style={yshift=-1mm, xshift=-0mm},
        ylabel={shift in steps},
        legend columns = 1,
        legend style={at={(axis cs: 0.,3.2)},anchor=north west, font = \small, legend cell align=center, align=center, draw=white!15!black,nodes={scale=0.75, transform shape}}
    ]

    \input{img/raw/exp/steps_low_error}
    
    \end{groupplot}
    
\end{tikzpicture}%
    }\\
    \subfloat[Ramp High Noise]{%
       \begin{tikzpicture}[baseline=(current bounding box.center)]
    
    \colorlet{sensor-rotation-0Color}{matlabRed!50!white}
    \colorlet{ground-truthColor}{matlabDarkRed}
    \colorlet{sensor-rotation-1Color}{matlabBlue!50!white}
    \colorlet{estimationColor}{matlabGreen}

    
    \colorlet{errorColor}{matlabRed}
    \colorlet{uncertaintyColor}{matlabBlue}

    \begin{groupplot}[
        group style={
            group size=1 by 2,
            group name=ramp,
            x descriptions at=edge bottom,
            y descriptions at=edge right,
            vertical sep=2.5pt,
            xlabels at=edge bottom
        },
        legend cell align={left},
        xlabel style={font=\color{white!15!black}, font=\small},
        xlabel={Time Steps $k$},
        minor x tick num=2,
        width=0.5 * 0.75\linewidth,
        scale only axis,
        xmin=0,
        xmax=194,
        xminorticks=true,
        axis background/.style={fill=white},
        xmajorgrids,
        xminorgrids,
        ymajorgrids,
        yminorgrids,
        major grid style={black!50},
        minor grid style={black!10},
        ticklabel style = {font=\small},
    ]
    
    \nextgroupplot[
        height = 0.5 * 0.4\linewidth,
        ymin=-0.05,
        ymax=1.25,
        yminorticks=true,
        ylabel style={yshift=2mm, xshift=-0mm},
        ylabel={rot. in $\si{\radian} / \text{step}$},
        legend columns = 1,
        legend style={at={(axis cs: 194,1.25)},anchor=north east, font = \small, legend cell align=center, align=center, draw=white!15!black,nodes={scale=0.75, transform shape}}
    ]

    \input{img/raw/exp/ramp_high_rotation}

    \nextgroupplot[
        height = 0.5 * 0.25\linewidth,
        ymin=-0.0,
        ymax=36,
        yminorticks=true,
        ytick={0, 12, 24},
        ylabel style={yshift=1mm, xshift=-0mm},
        ylabel={rot. change},
        legend columns = 1,
        legend style={at={(axis cs: 194.,36.)},anchor=north east, font = \small, legend cell align=center, align=center, draw=white!15!black,nodes={scale=0.75, transform shape}}
    ]

    \input{img/raw/exp/ramp_high_uncertainty}
    
    \end{groupplot}

    \begin{groupplot}[
        group style={
            group size=1 by 2,
            group name=ramp,
            x descriptions at=edge bottom,
            y descriptions at=edge left,
            vertical sep=2.5pt,
            xlabels at=edge bottom
        },
        legend cell align={left},
        minor x tick num=2,
        width=0.5 * 0.75\linewidth,
        scale only axis,
        xmin=0,
        xmax=194,
        ticklabel style = {font=\small},
    ]
    
    \nextgroupplot[
        height = 0.5 * 0.4\linewidth,
        ymin=-0.1,
        ymax=2.5,
        yminorticks=true,
        ylabel style={yshift=1mm, xshift=-0mm},
        ylabel={shift in steps},
        legend columns = 1,
        legend style={at={(axis cs: 0.,2.5)},anchor=north west, font = \small, legend cell align=center, align=center, draw=white!15!black,nodes={scale=0.75, transform shape}}
    ]

    \input{img/raw/exp/ramp_high_estimation}

    \nextgroupplot[
        height = 0.5 * 0.25\linewidth,
        ymin=-0.0,
        ymax=0.15,
        yminorticks=true,
        ytick={0, 0.1},
        ylabel style={yshift=-1mm, xshift=-0mm},
        ylabel={shift in steps},
        legend columns = 1,
        legend style={at={(axis cs: 0.,0.15)},anchor=north west, font = \small, legend cell align=center, align=center, draw=white!15!black,nodes={scale=0.75, transform shape}}
    ]

    \input{img/raw/exp/ramp_high_error}
    
    \end{groupplot}
    
\end{tikzpicture}%
    }
    \subfloat[Steps High Noise]{%
       \begin{tikzpicture}[baseline=(current bounding box.center)]
    
    \colorlet{sensor-rotation-0Color}{matlabRed!50!white}
    \colorlet{ground-truthColor}{matlabDarkRed}
    \colorlet{sensor-rotation-1Color}{matlabBlue!50!white}
    \colorlet{estimationColor}{matlabGreen}
    
    \colorlet{errorColor}{matlabRed}
    \colorlet{uncertaintyColor}{matlabBlue}

    \begin{groupplot}[
        group style={
            group size=1 by 2,
            group name=steps,
            x descriptions at=edge bottom,
            y descriptions at=edge right,
            vertical sep=2.5pt,
            xlabels at=edge bottom
        },
        legend cell align={left},
        xlabel style={font=\color{white!15!black}, font=\small},
        xlabel={Time Steps $k$},
        minor x tick num=2,
        width=0.5 * 0.75\linewidth,
        scale only axis,
        xmin=0,
        xmax=194,
        xminorticks=true,
        axis background/.style={fill=white},
        xmajorgrids,
        xminorgrids,
        ymajorgrids,
        yminorgrids,
        major grid style={black!50},
        minor grid style={black!10},
        ticklabel style = {font=\small},
    ]
    
    \nextgroupplot[
        height = 0.5 * 0.4\linewidth,
        ymin=-0.05,
        ymax=1.25,
        yminorticks=true,
        ylabel style={yshift=2mm, xshift=-0mm},
        ylabel={rot. in $\si{\radian} / \text{step}$},
        legend columns = 1,
        legend style={at={(axis cs: 194,1.25)},anchor=north east, font = \small, legend cell align=center, align=center, draw=white!15!black,nodes={scale=0.75, transform shape}}
    ]

    \input{img/raw/exp/steps_high_rotation}

    \nextgroupplot[
        height = 0.5 * 0.25\linewidth,
        ymin=-0.0,
        ymax=33,
        yminorticks=true,
        ytick={0, 11, 22},
        ylabel style={yshift=1mm, xshift=-0mm},
        ylabel={rot. change},
        legend columns = 1,
        legend style={at={(axis cs: 194.,33.)},anchor=north east, font = \small, legend cell align=center, align=center, draw=white!15!black,nodes={scale=0.75, transform shape}}
    ]

    \input{img/raw/exp/steps_high_uncertainty}
    
    \end{groupplot}

    \begin{groupplot}[
        group style={
            group size=1 by 2,
            group name=steps,
            x descriptions at=edge bottom,
            y descriptions at=edge left,
            vertical sep=2.5pt,
            xlabels at=edge bottom
        },
        legend cell align={left},
        width=0.5 * 0.75\linewidth,
        scale only axis,
        xmin=0,
        xmax=194,
        ticklabel style = {font=\small},
    ]
    
    \nextgroupplot[
        height = 0.5 * 0.4\linewidth,
        ymin=-0.1,
        ymax=3.75,
        yminorticks=true,
        ylabel style={yshift=1mm, xshift=-0mm},
        ylabel={shift in steps},
        legend columns = 1,
        legend style={at={(axis cs: 0.,3.75)},anchor=north west, font = \small, legend cell align=center, align=center, draw=white!15!black,nodes={scale=0.75, transform shape}}
    ]

    \input{img/raw/exp/steps_high_estimation}

    \nextgroupplot[
        height = 0.5 * 0.25\linewidth,
        ymin=-0.0,
        ymax=3.3,
        yminorticks=true,
        ytick={0, 1.1, 2.2},
        ylabel style={yshift=-1mm, xshift=-0mm},
        ylabel={shift in steps},
        legend columns = 1,
        legend style={at={(axis cs: 0.,3.3)},anchor=north west, font = \small, legend cell align=center, align=center, draw=white!15!black,nodes={scale=0.75, transform shape}}
    ]

    \input{img/raw/exp/steps_high_error}
    
    \end{groupplot}
    
\end{tikzpicture}%
    }
    \caption{%
        Results of different experiment settings are illustrated.
        For both ramp and steps error profiles (left/right) and two noise levels (top/bottom) the results are illustrated in the respective figure.
        In all figures, the ground truth time shift (red) and the estimated time shift (green) are plotted in the upper diagram.
        As a reference, the sensor rotation of the first simulation run is also shown in the background.
        For the estimated time shift, the green line represents the median value.
        The 25th and 75th quantiles are illustrated in gray.
        In the lower diagram, the absolute estimation error (red) and the estimation uncertainty (blue) are depicted.
    }
    \label{fig:exp_estimation}
\end{figure*}

We evaluated our offset estimation approach on two different error types.
A ramp-shaped and a step error profile is used as a time offset.
In both cases, the initial time offset between both sensors is zero, meaning that they are in sync.
Then, in the ramp-shape case, the time offset slowly increases over time, whereas, in the step case, the time offset changes rapidly between two consecutive time steps.
Data was simulated for both error types using two different noise levels, namely, $\SI{50}{\percent}$, called low noise, and $\SI{200}{\percent}$, called high noise in the following.
It shall be noted that the noise values are with respect to the fine simulation grid.
The results are illustrated in Fig.~\ref{fig:exp_estimation}.

Generally, our time offset approach is able to estimate both types of error profiles.
As expected, the results of the low noise case are more precise than those of the high noise case.
However, even under the influence of high noise, the median error is low in most cases.
Given the ramp-shaped profile, our uncertainty strongly correlates with the error value.
Most importantly, whenever the uncertainty is low, the estimation error is also low.
Further, the 25th and 75th quantiles show that the estimation variance also correlates to the uncertainty measure.
It is also small whenever the uncertainty is low.
This shows that given slow variations are successfully detected by our time offset estimation.
For the step error profile, the results also show accurate results in most cases; however, whenever there is an offset change, the estimation requires some time to follow.
This can be explained by the fact that a change can only be detected if there is enough rotational change inside the sliding window.
Comparing the estimation errors and their uncertainty measures, it can be seen that when the uncertainty is low, the estimation error rapidly decreases, but in the case of higher uncertainty values during a step, the error is only gradually reduced over time. 
Similar to the ramp-shaped case, the estimation variance correlates with the uncertainty measure, which further demonstrates its practicability.
The single-threaded non-optimized median run time of our approach was $\SI{60}{\milli\second}$, which indicates online capable execution.

\subsection{Vehicle Tracking}
\label{sec:trackeval}
\begin{figure*}
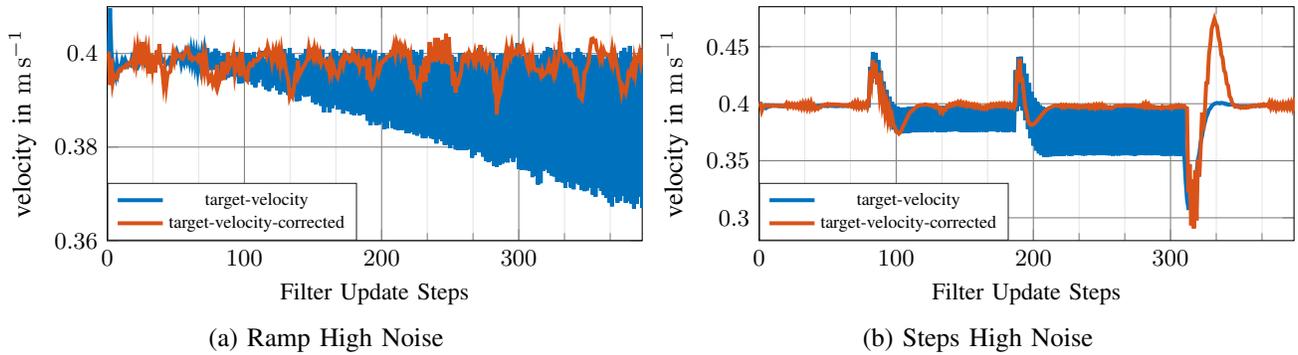

    \centering
    \captionsetup[subfloat]{labelfont=normalsize,textfont=normalsize}
    \subfloat[Ramp High Noise]{%
       \begin{tikzpicture}[baseline=(current bounding box.center)]
    
    \colorlet{target-velocityColor}{matlabBlue}
    \colorlet{target-velocity-correctedColor}{matlabRed}

    \begin{groupplot}[
        group style={
            group size=1 by 1,
            group name=ramp,
            x descriptions at=edge bottom,
            y descriptions at=edge left,
            vertical sep=2.5pt,
            xlabels at=edge bottom
        },
        legend cell align={left},
        xlabel style={font=\color{white!15!black}, font=\small},
        xlabel={Filter Update Steps},
        minor x tick num=2,
        width=0.5 * 0.8\linewidth,
        scale only axis,
        xmin=0,
        xmax=390,
        xminorticks=true,
        axis background/.style={fill=white},
        xmajorgrids,
        xminorgrids,
        ymajorgrids,
        yminorgrids,
        major grid style={black!50},
        minor grid style={black!10},
        ticklabel style = {font=\small},
    ]
    
    \nextgroupplot[
        height = 0.5 * 0.35\linewidth,
        ymin=0.36,
        ymax=0.41,
        yminorticks=true,
        ylabel style={yshift=0mm, xshift=-0mm},
        ylabel={velocity in $\si{\metre\per\second}$},
        legend columns = 1,
        legend style={at={(axis cs: 0,0.36)},anchor=south west, font = \small, legend cell align=center, align=center, draw=white!15!black,nodes={scale=0.75, transform shape}}
    ]

    \input{img/raw/exp/ramp_high_filter}

    \end{groupplot}
    
\end{tikzpicture}%
    }
    \subfloat[Steps High Noise]{%
       \begin{tikzpicture}[baseline=(current bounding box.center)]
    
    \colorlet{target-velocityColor}{matlabBlue}
    \colorlet{target-velocity-correctedColor}{matlabRed}

    \begin{groupplot}[
        group style={
            group size=1 by 1,
            group name=ramp,
            x descriptions at=edge bottom,
            y descriptions at=edge left,
            vertical sep=2.5pt,
            xlabels at=edge bottom
        },
        legend cell align={left},
        xlabel style={font=\color{white!15!black}, font=\small},
        xlabel={Filter Update Steps},
        minor x tick num=2,
        width=0.5 * 0.8\linewidth,
        scale only axis,
        xmin=0,
        xmax=390,
        xminorticks=true,
        axis background/.style={fill=white},
        xmajorgrids,
        xminorgrids,
        ymajorgrids,
        yminorgrids,
        major grid style={black!50},
        minor grid style={black!10},
        ticklabel style = {font=\small},
    ]
    
    \nextgroupplot[
        height = 0.5 * 0.35\linewidth,
        ymin=0.28,
        ymax=0.485,
        yminorticks=true,
        ylabel style={yshift=0mm, xshift=-0mm},
        ylabel={velocity in $\si{\metre\per\second}$},
        legend columns = 1,
        legend style={at={(axis cs: 0,0.28)},anchor=south west, font = \small, legend cell align=center, align=center, draw=white!15!black,nodes={scale=0.75, transform shape}}
    ]

    \input{img/raw/exp/steps_high_filter}

    \end{groupplot}
    
\end{tikzpicture}%
    }
    \caption{%
        Results of the velocity estimation using a Kalman filter are illustrated.
        Each figure shows the results for the respective scenario of Fig.~\ref{fig:exp_estimation}.
        Thereby, the velocity estimation without any correction (blue) and the velocity estimation after applying our approach (red) are shown.
        The actual velocity of the vehicle is $\SI{0.4}{\metre\per\second}$.
    }
    \label{fig:exp_filter}
\end{figure*}
%
In this section, the applicability of our approach is demonstrated.
For this, sensor measurements of a target vehicle have been simulated as described earlier in Section~\ref{sec:dataSim}.
Thereby, the target vehicle follows a straight course with a constant velocity, and each sensor measures its position.
While the ego vehicle moves in two dimensions, the target vehicle only moves along the x-axis for simplicity and demonstration reasons; however, the experiments of this section do not require any specific motion direction.
Further, a Kalman filter is used to track the target vehicle.
Since the motion model is linear and zero-mean, white Gaussian noise is used during the simulation; a Kalman filter is optimal.
This will emphasize the importance of synchronization even under near-perfect conditions, i.e., optimal estimation with known motion models and noise characteristics.

Given two synchronized sensors, the expected output of a Kalman filter using the measurements of two sensors is an accurate vehicle track.
Thus, in this section, we assume that any inconsistency in tracking results is mostly due to synchronization offsets.
Using a measurement noise standard deviation $\sigma_r = \SI{5}{\centi\metre}$ and a process noise standard deviation of $\sigma_q = \SI{1}{\milli\metre} \cdot dt$ the median velocity estimation results of the Kalman filters are illustrated in Fig.~\ref{fig:exp_filter}.
In each simulation run, all measurements were ordered by their time and then fed to the Kalman filter.
Thus, twice as many sample points are available.
By showing the filter results after each measurement instead of in a given time grid, the influence of time offsets is better visible.

The results show that the time offset compromises the filter performance noticeably in both error patterns.
Without our correction approach, the filter output contains bigger jumps, becoming inconsistent when the offset increases.
In contrast, using the estimated offset as correction reduces the impact.
Especially in the case of step errors, the filter performance is significantly better after an initial oscillation phase.
When there is no error, any misdetected time offset reduces the filter performance, which suggests that a correction strategy should not alter measurement times under a certain threshold.

Generally, this section's results show that our approach reduces the impact of sensor synchronization inconsistencies in tracking systems.
Even in the simplistic scenario, time offsets lead to significant performance losses.

\section{Conclusion}
\label{sec:conclusion}
In this work, we first derived an approach to estimate time-dependent, changing time offsets between two sensors together with an uncertainty measure of the estimation result.
Further, a self-assessment of sensor synchronicity and different, task-specific correction strategies have been described. 
Our method was evaluated using Monte Carlo simulation, showing accurate estimation results.
Based on the uncertainty measure, the self-assessment reliably assesses the estimation quality, i.e., it is strongly correlated with the estimation variance and error.
Investigating the effect of synchronization inconsistencies on tracking systems underlined the importance of measurement synchronization.
By applying our approach, most negative impacts could be mitigated, proving the applicability of our proposed method.
For future work, it seems promising to integrate the approach in multi-sensor autonomous systems, allowing advanced correction strategies and real-world experiments.

\bibliographystyle{IEEEtran}
\bibliography{IEEEabrv,references}

\begin{thebibliography}{10}
\providecommand{\url}[1]{#1}
\csname url@rmstyle\endcsname
\providecommand{\newblock}{\relax}
\providecommand{\bibinfo}[2]{#2}
\providecommand\BIBentrySTDinterwordspacing{\spaceskip=0pt\relax}
\providecommand\BIBentryALTinterwordstretchfactor{4}
\providecommand\BIBentryALTinterwordspacing{\spaceskip=\fontdimen2\font plus
\BIBentryALTinterwordstretchfactor\fontdimen3\font minus \fontdimen4\font\relax}
\providecommand\BIBforeignlanguage[2]{{%
\expandafter\ifx\csname l@#1\endcsname\relax
\typeout{** WARNING: IEEEtran.bst: No hyphenation pattern has been}%
\typeout{** loaded for the language `#1'. Using the pattern for}%
\typeout{** the default language instead.}%
\else
\language=\csname l@#1\endcsname
\fi
#2}}

\bibitem{jellum2024sync}
E.~R. Jellum, T.~H. Bryne, T.~A. Johansen, and M.~Orland{\'{i}}c, ``{The Syncline Model – Analyzing the Impact of Time Synchronization in Sensor Fusion},'' 2024.

\bibitem{ieee1588}
``{IEEE Standard for a Precision Clock Synchronization Protocol for Networked Measurement and Control Systems},'' \emph{IEEE Std 1588-2008 (Revision of IEEE Std 1588-2002)}, pp. 1--269, 2008.

\bibitem{iso21448}
{International Organization for Standardization}, ``{ISO/PAS 21448: Road Vehicles –— Safety of the Intended Functionality},'' \emph{ISO, Publicly Available Specification}, 2019.

\bibitem{griebel2023online}
T.~Griebel, J.~Heinzler, M.~Buchholz, and K.~Dietmayer, ``{Online Performance Assessment of Multi-Sensor Kalman Filters Based on Subjective Logic},'' in \emph{International Conference on Information Fusion (FUSION)}, 2023.

\bibitem{daniilidis1999handeye}
\BIBentryALTinterwordspacing
K.~Daniilidis, ``{Hand-Eye Calibration Using Dual Quaternions},'' \emph{The International Journal of Robotics Research}, vol.~18, no.~3, pp. 286--298, 1999. [Online]. Available: \url{https://doi.org/10.1177/02783649922066213}
\BIBentrySTDinterwordspacing

\bibitem{horn2021online}
M.~Horn, T.~Wodtko, M.~Buchholz, and K.~Dietmayer, ``{Online extrinsic calibration based on per-sensor ego-motion using dual quaternions},'' \emph{IEEE Robotics and Automation Letters}, vol.~6, no.~2, 2021.

\bibitem{taylor2016motion}
Z.~Taylor and J.~Nieto, ``{Motion-Based Calibration of Multimodal Sensor Extrinsics and Timing Offset Estimation},'' \emph{IEEE Transactions on Robotics}, vol.~32, no.~5, pp. 1215--1229, 2016.

\bibitem{furrer2017Evaluation}
F.~Furrer, M.~Fehr, T.~Novkovic, H.~Sommer, I.~Gilitschenski, and R.~Siegwart, ``{Evaluation of Combined Time-Offset Estimation and Hand-Eye Calibration on Robotic Datasets},'' 3 2017.

\bibitem{excalibur}
\BIBentryALTinterwordspacing
M.~Horn and T.~Wodtko, ``{Excalibur: An open-source Python library for extrinsic sensor calibration},'' 2023. [Online]. Available: \url{https://github.com/uulm-mrm/excalibur}
\BIBentrySTDinterwordspacing

\bibitem{bracewell1996fourier}
\BIBentryALTinterwordspacing
R.~N. Bracewell, ``{The Fourier Transform and Its Applications},'' 1966. [Online]. Available: \url{https://api.semanticscholar.org/CorpusID:18010056}
\BIBentrySTDinterwordspacing

\bibitem{kalman1960new}
R.~E. Kalman, ``{A New Approach to Linear Filtering and Prediction Problems},'' \emph{Journal of Basic Engineering}, vol.~82, no.~1, pp. 35--45, 3 1960.

\end{thebibliography}

\end{document}